\documentclass{article}

\usepackage{microtype}
\usepackage{graphicx}
\usepackage{subfigure}
\usepackage{booktabs} 

\usepackage{hyperref}

\usepackage[accepted]{icml2025}

\usepackage{amsmath}
\usepackage{amssymb}
\usepackage{mathtools}
\usepackage{amsthm}

\usepackage{soul}

\usepackage{amsmath}
\usepackage{amssymb}
\usepackage{amsfonts}
\usepackage{mathtools}
\usepackage{amsthm}
\usepackage{bm}
\usepackage{mathtools}
\usepackage{thm-restate}
\usepackage{multirow}  
\usepackage{dsfont}
\usepackage{enumitem}
\usepackage{multirow}
\usepackage{indentfirst}
\usepackage{wrapfig}
\usepackage{pifont}
\usepackage{booktabs}
\usepackage{upgreek}

\usepackage{graphicx} 
\usepackage{url}
\usepackage{hyperref}
\usepackage{blindtext}
\usepackage{parskip}
\usepackage{color}
\usepackage[framemethod=TikZ]{mdframed}
\usepackage{tikz}
\usepackage{mathabx}
\usepackage[many]{tcolorbox}    	

\usepackage{microtype}

\usepackage[capitalize,noabbrev]{cleveref}

\usepackage[textsize=tiny]{todonotes}

\usepackage[framemethod=TikZ]{mdframed}
\mdfdefinestyle{contribution}{%
    middlelinecolor =   none,
    middlelinewidth =   1pt,
    backgroundcolor =   mygreen!30,
    roundcorner     =   5pt,
}
\mdfdefinestyle{takeaway}{%
    middlelinecolor =   none,
    middlelinewidth =   1pt,
    backgroundcolor =   mygreen!10,
    roundcorner     =   5pt,
}

\definecolor{main}{HTML}{5989cf}    
\definecolor{sub}{HTML}{cde4ff}     
\definecolor{text}{HTML}{3467B4}

\newtcolorbox{mybox}{
    sharpish corners, 
    colback = sub, 
    colframe = main, 
    boxrule = 0pt, 
    toprule = 4.5pt, 
    enhanced,
    left = 1pt,
    right = 1pt,
    fuzzy shadow = {0pt}{-2pt}{-0.5pt}{0.5pt}{black!35} 
}

\usepackage{transparent}

\usepackage{hyperref}
\hypersetup{
    colorlinks=true,
    linkcolor=orange,
    filecolor=magenta,      
    urlcolor=cyan,
    citecolor=blue,
    pdftitle={eri},
    pdfpagemode=FullScreen,
    }

\usepackage[capitalize,noabbrev]{cleveref}   
\usepackage{lipsum}

\newcommand*\circled[1]{\tikz[baseline=(char.base)]{
            \node[shape=circle,draw,inner sep=1pt] (char) {#1};}}
\newcommand{\Exp}[2]{\mathbb{E}_{#1}\left[#2\right]}
\newcommand{\cmark}{\ding{51}}%
\newcommand{\xmark}{\ding{55}}%

\newcommand{\bmag}[1]{\textbf{\textcolor{text}{#1}}}

\definecolor{mypurple}{RGB}{217, 178, 235}
\definecolor{mygreen}{RGB}{0, 179, 0}

\theoremstyle{plain}
\newtheorem{theorem}{Theorem}[section]

\theoremstyle{definition}

\newtheorem{assumption}[theorem]{Assumption}
\newtheorem{remark}[theorem]{Remark}

\usepackage[capitalize,noabbrev]{cleveref}

\usepackage[textsize=tiny]{todonotes}

\icmltitlerunning{Beyond the ATE: Interpretable Modelling of Treatment Effects over Dose and Time}

\begin{document}

\twocolumn[
\icmltitle{Beyond the ATE: Interpretable Modelling of Treatment Effects \\ over Dose and Time}



\icmlsetsymbol{equal}{*}

\begin{icmlauthorlist}
\icmlauthor{Julianna Piskorz}{cam}
\icmlauthor{Krzysztof Kacprzyk}{cam}
\icmlauthor{Harry Amad}{cam}
\icmlauthor{Mihaela van der Schaar}{cam}
\end{icmlauthorlist}

\icmlaffiliation{cam}{Department of Mathematics and Theoretical Physics, University of Cambridge, United Kingdom}

\icmlcorrespondingauthor{Julianna Piskorz}{jp2048@cam.ac.uk}

\icmlkeywords{Machine Learning, Interpretability, Causal Inference}

\vskip 0.3in
]



\printAffiliationsAndNotice{}  

\begin{abstract}
The Average Treatment Effect (ATE) is a foundational metric in causal inference, widely used to assess intervention efficacy in randomized controlled trials (RCTs). However, in many applications -- particularly in healthcare -- this static summary fails to capture the nuanced dynamics of treatment effects that vary with both dose and time. We propose a framework for modelling treatment effect trajectories as smooth surfaces over dose and time, enabling the extraction of clinically actionable insights such as onset time, peak effect, and duration of benefit. To ensure interpretability, robustness, and verifiability -- key requirements in high-stakes domains -- we adapt \texttt{SemanticODE}, a recent framework for interpretable trajectory modelling, to the causal setting where treatment effects are never directly observed. Our approach decouples the estimation of trajectory shape from the specification of clinically relevant properties (e.g., maxima, inflection points), supporting domain-informed priors, post-hoc editing, and transparent analysis. We show that our method yields accurate, interpretable, and editable models of treatment dynamics, facilitating both rigorous causal analysis and practical decision-making.
\end{abstract}

\section{Introduction}
\label{sec: intro}

\textbf{Average effects are the backbone of causal inference.} Randomized controlled trials (RCTs) are the gold standard for evaluating the efficacy of treatments, with the \textit{Average Treatment Effect} (ATE) serving as their foundational metric. The ATE provides a straightforward and robust measure, allowing to determine whether, on average, a treatment positively influences outcomes within a specified population. Its simplicity not only facilitates rigorous verification and validation but also establishes trustworthiness, making the ATE a cornerstone in the analysis of RCTs.

Despite its strengths, the ATE for binary treatments offers only a \textit{single summary statistic} and thus cannot address more nuanced or clinically relevant questions -- such as determining the optimal timing and dosage of treatments, or scheduling follow-up appointments to maximize patient outcomes. When treatments allow for different dose levels, the \textit{Average Dose Response Function} (ADRF) \citep{hirano_propensity_2004} extends the ATE by providing a one-dimensional (and still transparent) mapping from the dose to expected outcome. While the ADRF provides more flexibility than standard ATE, it still captures effects at a \textit{single point in time}, overlooking the dynamic nature of treatment effects as they evolve. This temporal aspect is crucial for guiding not just the decision of whether to administer a treatment, but also \textit{when} and \textit{how} to do so to achieve maximum efficacy.

\begin{figure*}[th]
    \centering
    \includegraphics[width=\linewidth]{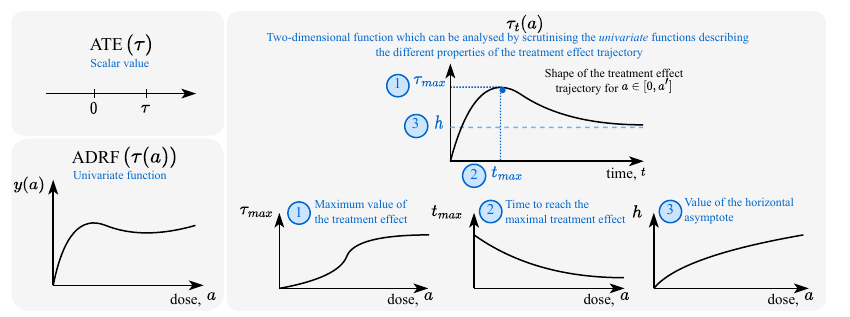}
    \caption{Despite being a two-dimensional function, $\tau_t(a)$ can be visualised and analysed by scrutinising the univariate functions which describe the dependence of its key properties on the dose of the treatment assigned.}
    \label{fig: fig1}
\end{figure*}

\textbf{Can we add the temporal dimension to the analysis of RCTs, without sacrificing interpretability?} To allow for more actionable insights, which can drive data-informed decision making, in this work our goal is to model the average treatment effect as a smooth, continuous function of both dose and time. Doing so allows us to answer the following questions, which are not approachable using static summaries (such as ADRF): $\blacktriangleright$ \textit{How long does it typically take for the treatment effect to exceed a clinically relevant threshold $\alpha$? $\blacktriangleright$ When, on average, does the treatment effect reach its peak, and how does this vary with the dose? $\blacktriangleright$ What is the lowest dose ensuring that the effect has a sustained benefit?} These insights are vital in clinical practice, supporting precise dosage adjustments, thorough (longitudinal) assessment of treatment efficacy, and strategic scheduling of follow-up appointments to ensure patient safety and optimal outcomes \citep{han_finding_2018}.

Crucially, recognising the importance of verifiability and interpretability in high-stakes domains, we aim to obtain dose-time treatment effect models which achieve high accuracy, while remaining transparent, editable, and easy to analyse. However, the two-dimensional nature of the treatment effect surface that we consider introduces significant challenges in interpretability, particularly compared to the simpler ATE or ADRF (\cref{fig: fig1}), since visualizing a function of two continuous variables is challenging. Furthermore, to ensure robustness, domains such as healthcare require solutions which can easily incorporate prior knowledge and requirements (e.g., the treatment effect should be zero initially) -- a requirement which existing modelling methods rarely satisfy.

\textbf{Our solution.} To satisfy the above requirements of interpretability and robustness, we leverage \texttt{SemanticODE} \citep{kacprzyk_no_2024}, a recently proposed framework designed for interpretable modelling of temporal trajectories. SemanticODE first fits a function that captures the general shape of the trajectory, known as the \emph{composition}, and then separately models the \emph{properties} of each composition, such as minima, maxima, and inflection points. This modular approach enhances transparency and allows domain experts to verify that the model's predictions are consistent with established domain knowledge, also offering a straightforward way to edit the final model, if necessary. In this work, we adapt SemanticODE to the problem of treatment effect estimation, where the target trajectory -- the treatment effect trajectory -- is never directly observed in the data, and the shape of the trajectory depends on the value of the dose, $a$, rather than on some initial condition.

\textbf{Contributions and outline.} In \cref{sec: setting} we formally introduce the setting considered in the work and explain how modelling treatment effects as a function of dose and time yields clinically actionable insights. In \cref{sec: estimation} we discuss the estimation strategies which can be used to ensure that the obtained model of the treatment effect is not only interpretable, but also unbiased. \cref{sec: semantic} provides an overview of SemanticODE and details the adaptations required to adjust it to the treatment effect setting. Finally, in \cref{sec: experiments}, we empirically demonstrate that our approach enables accurate predictions and offers interpretable, verifiable, and editable summaries of treatment dynamics.

\section{Related Works}
\label{sec: related}

\paragraph{Time-varying causal inference.} Most of existing causal estimands either capture limited temporal insights or present such complexity that verification and thorough analysis become impractical. Below, we briefly survey two main categories of time-varying causal estimands, with further details and comparison in \cref{appendix: related}.

\textit{Sequential ATE.} Early works on time-varying treatments focus on the \emph{average treatment effect} of a sequence of categorical treatments \citep{robins_new_1986, robins_correcting_1994, robins_marginal_2000}. Although straightforward to interpret, these methods analyse the cumulative effects of treatments only at fixed time points, offering limited insight into how treatment effects (smoothly) \emph{evolve} over time or how this evolution depends on the dose.

\textit{Sequential CATE.} Drawing on the sequential ATE setting, multiple works also study \emph{personalised} (conditional average) effects, modelling outcome trajectories based on evolving covariate and treatment histories, both in discrete-time \citep{bica_estimating_2019, li_g-net_2021, melnychuk_causal_2022} and continuous-time \citep{seedat_continuous-time_2022, hess_bayesian_2024, kacprzyk_ode_2024} settings. While these approaches indeed model the treatment effect dynamics, and can yield accurate predictions for specific combinations of treatment sequences and patient histories, the complexity of exploring all possible combinations of inputs makes them challenging to validate in high-stakes domains. It also hinders crucial questions for model understanding \citep{liao_questioning_2020}, such as how specific covariates might alter the outcome or how to modify covariates to achieve different predictions. In contrast to this body of works, in this work we decide to focus on estimating the \textit{average} (rather than personalised) treatment effect, of a \textit{single treatment} (rather than sequences of treatments). As we demonstrate, such an approach allows to obtain interpretable and verifiable solutions.
\section{Problem Setting}
\label{sec: setting}

In this work, we consider a problem setting where each patient can be assigned a dose $A \in \mathcal{A} = \{0\} \cup [a_{min}, a_{max}]$ of a treatment, with $a=0$ representing the baseline condition of \textit{no treatment}. Our objective is to evaluate the efficacy of the treatment at various time points $T \in  \mathcal{T} = [0, t_{max}]$, where we assume that the treatment was administered at the time $t=0$. To quantify the effectiveness of the treatment, we introduce an outcome variable $Y \in \mathcal{Y} \subseteq \mathbb{R}$, which for each patient can be measured at multiple time points $t$, and denote its value at time $T=t$ as $Y_t$.

Since at any time $t > 0$ we only observe the outcome $Y_t$ associated with the actually administered treatment dose $A$ (commonly referred to as the \textit{factual} outcome), following Rubin's potential outcome framework and its extensions to continuous treatments \citep{rubin_bayesianly_1984, hirano_propensity_2004}, we assume that each patient is characterised by a possibly infinite set of potential outcomes $Y_t(a)$ associated with other feasible treatment doses $a$. The \emph{observed} outcome is then defined by $Y = Y_T(A)$, where $A$ is the assigned dose and $T$ is the chosen time of measurement. With this notation, we can now formally define our target estimand:

\begin{equation}
\tau_t(a) = \Exp{}{Y_t(a) - Y_t(0)}.
\end{equation}

Let us now summarise the key characteristics of $\tau_t(a)$, as well as their medical relevance and significance:

\begin{itemize}
\item \bmag{Point treatment:} We focus on point treatments administered at a single time point ($t=0$) and remaining fixed throughout the study. This setting applies to one-off interventions (e.g., medical procedures, single drug injections) and long-term treatments that are assigned once and then administered regularly without modification (e.g., chronic disease medication).

\item \bmag{Continuous treatment:} Furthermore, $\tau_t(a)$ allows for a continuous treatment variable, the magnitude of which can represent factors such as the dose of medication or duration of exposure (e.g., oxygen therapy).

\item \bmag{Focus on the treatment effect:} Rather than modelling the average potential outcome surfaces, $\mathbb{E}[Y_t(a)]$, by estimating $\tau_t(a)$ we focus on the \textit{treatment effect} directly. This approach offers several advantages: it enables the imposition of clear inductive biases (e.g., the treatment effect should be zero at $t=0$ or -- for some treatments -- should decay to zero over time), facilitates straightforward comparison of the effect of different doses, and, as the difference of two potentially complex functions, it can exhibit simpler structures that lend themselves to transparent modelling \cite{curth_inductive_2021}.\footnote{While in this work we focus on the more complex task of the estimation of the treatment effect, the results and methods presented in this work can be easily extended to transparent modelling of $\mathbb{E}[Y_t(a)]$ instead, if that choice was more aligned with the requirements of a given end user.}

\item \bmag{Average treatment effect:} we focus on the population-level average treatment effect, rather than on the personalised predictions. This allows us to use transparent models which are amenable to expert verification.
\end{itemize}

\subsection{Why Estimate Treatment Effects as a Smooth Function of Time and Dose?}
\label{sec: insights}
With $\tau_t(a)$, we aim to capture the evolution of the average treatment effect over time across different dose levels, offering actionable insights that go beyond simpler \textit{static} measures such as the ATE or the ADRF. Functional analysis of the estimated $\tau_t(a)$ allows to obtain the following insights:

\textbf{Identifying optimal dosage.} Unlike static causal estimands, $\tau_t(a)$ enables more comprehensive dose assignment decisions \citep{han_finding_2018} by allowing insights into factors such as the time required to achieve maximal effect (by finding the coordinates of the maximum point of $\tau_t(a)$), or the duration for which the treatment effect remains above a clinically significant threshold.

\textbf{Insights into expected response.} $\tau_t(a)$ also provides deeper insights into the expected treatment response under a fixed dose. For example, it allows to predict the average time needed for the treatment effect to reach a clinically meaningful level, which can inform decisions on when to introduce alternative or backup treatments. Additionally, by allowing to predict when the treatment effect is likely to decline below a critical level, modelling $\tau_t(a)$ can help clinicians schedule follow-up visits more effectively, ensuring sustained therapeutic benefits.

These insights, grounded in modelling the treatment effect as continuous functions of both dose and time, provide a more comprehensive foundation for data-driven decision-making in high-stakes domains such as healthcare.

\subsection{$\tau_t(a)$ Facilitates Transparent and Verifiable Solutions}
\label{sec: dalte_transparent}

By incorporating the temporal dimension to the ATE, $\tau_t(a)$ provides more actionable insights compared to simpler measures such as the ATE or ADRF. However, a key advantage of these traditional measures is their simplicity, which allows for robust scrutiny by domain experts. ATE, which provides a single scalar value summarizing the overall effect, and ADRF, which offers a one-dimensional dose-response relationship, can both be easily visualized and interpreted across the entire domain, making them reliable and widely accepted tools for decision-making.

Despite the additional complexity introduced by the two-dimensional nature of $\tau_t(a)$ -- where each dose level corresponds to a distinct time-varying treatment effect trajectory -- we argue that a model for estimating $\tau_t(a)$ can remain transparent and verifiable. While directly visualizing and analysing the treatment effect trajectory at all dose levels may be impractical (if not impossible), the key properties necessary to unlock the actionable insights (as discussed in \cref{sec: insights}) can be effectively summarized using interpretable \textit{univariate} functions of the dose. Specifically, critical features of each time-varying trajectory, such as the maximum treatment effect value, the time required to reach this maximum, the value of the horizontal asymptote, provide meaningful insights into the model’s predictions. However, as they can be described as functions of the dose only, they can be easily visualised and verified, enabling domain experts to assess and interpret the model used to fit $\tau_t(a)$. A visual illustration of this concept is provided in \cref{fig: fig1}.

For example, by looking at the relationship between the maximum treatment effect and the dose (plot 1), domain experts can \textit{verify} whether the model agrees with known dose-response relationships (e.g. stronger dose leads to larger treatment effect). Further, by analysing the relationship between the time needed to reach the maximum effect and dose (plot 2), the expert can decide what is the smallest dose that will lead to a sufficiently fast response.

While not all methods for modelling $\tau_t(a)$ inherently support such analysis of the trajectory's properties, adopting principles of direct semantic modelling -- as discussed in \cref{sec: semantic} -- achieves both accuracy and transparency by learning these properties directly from the data and embedding them in the model design. This leads to interpretable-by-design solutions that align well with the goals of treatment effect estimation, ensuring verifiability and trustworthiness.

\section{Estimation Strategies}
\label{sec: estimation}
Having introduced our objective $\tau_t(a)$ and its significance in clinical applications, we now discuss potential estimation strategies.

In this preliminary work, we assume that the dataset used for the estimation of $\tau_t(a)$ originates from a randomized controlled trial (RCT), and can be represented as follows:
$$\mathcal{D} = \{\bm{X}_i, A_i, (T_{i, j}, Y_{i,j})_{j=1}^{m_i}\}_{i=1}^n,$$
where for each patient $i$, characterised by covariates $\bm{X}_i$ and assigned dose $A_i$, we observe $m_i$ repeated outcomes $Y_{i,j} \in \mathbb{R}$, measured at irregular time points $T_{i, j} \in \mathcal{T} \subseteq \mathbb{R}^+$. Thus, the outcome trajectory $(T_{i, j}, Y_{i,j})_{j=1}^{m_i}$ captures the temporal evolution of the outcome $Y$ for each patient.

\subsection{Identifiability Assumptions}
To ensure unbiased estimation of $\tau_t(a)$ using this dataset, we impose the following assumptions about the design of the RCT and the underlying data-generating process.

\begin{assumption}[Independent Treatment Assignment]
\label{assumption: independent}
    The treatment dose is \textit{randomly} assigned to the patients, ensuring that for any fixed value $T=t$, and for all $a \in \mathcal{A}$, $Y_t(a) \perp A$.
\end{assumption}

Even with random treatment assignment, the obtained estimates of $\tau_t(a)$ could still be biased due to informative sampling of outcomes. If the timing of measurements depends on the underlying patient characteristics -- such as evolving health status -- bias can arise because the subpopulations observed at different time points may differ systematically \citep{vanderschueren_accounting_2023, lin_analysis_2004, robins_analysis_1995}. To mitigate this potential source of bias, we make an additional assumption regarding the timing of measurements.

\begin{assumption}[Sampling Completely at Random]
\label{assumption: SCAR}
    For each patient, the timings of the outcome measurements are independent of the assigned dose or individual patient characteristics, ensuring that for any fixed value $T=t$, and for all $a \in \mathcal{A}$, $Y_t(a) \perp T$.
\end{assumption}

\begin{figure}[h]
    \centering
    \includegraphics[width=0.3\linewidth]{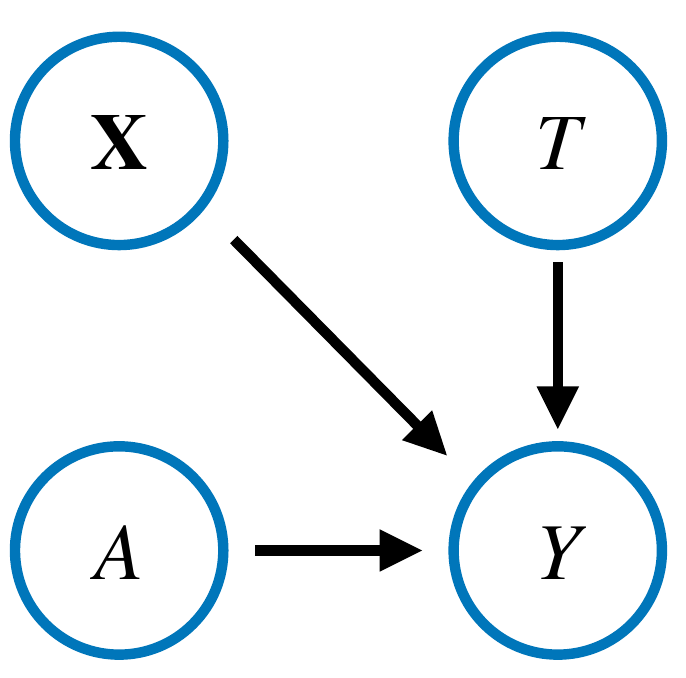}
    \caption{The causal graph characterising the RCT setting considered in this work.}
    \label{fig: causal_graph}
\end{figure}

The corresponding causal graph is illustrated in \cref{fig: causal_graph}. For completeness, we make the consistency assumption: $Y = Y_T(A)$ and the overlap assumptions: $P(A = a) > 0 \forall a \in \mathcal{A}$ and $P(T = t) > 0 \; \forall t \in \mathcal{T}$.

Given \cref{assumption: independent} and \cref{assumption: SCAR}, for fixed values of $A=a$ and $T=t$, there are no backdoor paths between the treatment dose, measurement time, and outcome. As a result, patients assigned different doses and measured at different time points remain exchangeable, allowing models trained on subsets of the population (conditioned on $A=a$ and $T=t$) to generalize to the entire population. Namely, the $\tau_t(a)$ function at any point $(a, t)$ can be estimated as:
\begin{align}
\label{eq: estimation}
    \tau_t(a) &= \mathbb{E}[Y_t(a) - Y_t(0)] \\
    &= \mathbb{E}[Y \vert A = a, T = t] - \mathbb{E}[Y \vert A = 0, T = t]
\end{align}
However, this only allows estimation for points $(a, t)$ that are directly observed in the dataset, significantly limiting the scope of inference. To enable meaningful estimation of $\tau_t(a)$ across all dose levels $a \in \mathcal{A}$ and all time points $t \in \mathcal{T}$, we further assume that the function $\tau_t(a)$ is \textit{sufficiently smooth} with respect to both $a$ and $t$, allowing for reliable interpolation and generalisation to unobserved values.

\vspace{1em}
\begin{remark}[Can we Make Weaker Assumptions?]
    While the assumptions imposed in this work ensure unbiased estimation of the $\tau_t(a)$ function, they could be relaxed to allow for a more flexible data-generating process. However, as further discuss in \cref{appendix: estimation}, estimation under these more relaxed assumptions would necessitate the development and fitting of more complex, high-dimensional models to adequately adjust for confounding and selection bias. As the focus of this work is to show the utility of $\tau_t(a)$ in RCT analysis, we opt for the stronger assumptions to facilitate transparent and interpretable analysis.
\end{remark}

\subsection{Ensuring Transparency in the $\tau_t(a)$ Model}
The implication of \cref{eq: estimation} is that $\tau_t(a)$ can be estimated from the dataset $\mathcal{D}$ by fitting a regression model $\hat{\mu}_t(a)$ to approximate the conditional expectation function $\mu_t(a) = \mathbb{E}[Y \vert A = a, T = t]$, while ignoring any patient-specific covariates $\bm{X}$. The estimate of $\tau_t(a)$ can then be computed as the difference between the treated and baseline outcomes: $\hat{\tau}_t(a) = \hat{\mu}_t(a) - \hat{\mu}_t(0)$.

Although under Assumptions \ref{assumption: independent} and \ref{assumption: SCAR} this approach ensures unbiased estimation of $\tau_t(a)$, it raises challenges related to \textit{transparency} and \textit{verifiability}. Specifically, the key obstacle is that \textit{the difference of two transparent models is not necessarily transparent}. Consequently, using a transparent model for $\mu_t(a)$ does not guarantee that the resulting estimate of $\tau_t(a)$ will retain the desired ease of verification, which is the primary objective of this work.
Additionally, since treatment effect trajectories $\tau_t(a)$ are often expected to exhibit simpler, smoother patterns compared to the potential outcome trajectories \citep{curth_inductive_2021}, imposing sparsity constraints on the more complex outcome functions $\mu_t(a)$, rather than directly on $\tau_t(a)$, may lead to suboptimal model performance.

To overcome these issues, we propose an alternative approach to the estimation of $\tau_t(a)$ that prioritises both transparency and accuracy. Instead of directly modelling the outcome function from the data, we propose to first construct individual \textit{surrogate treatment effects} $\tilde{\tau}_t(a)$, which capture the individual difference in potential outcomes. We then estimate $\tau_t(a)$ by regressing $\tilde{\tau}_t(a)$ on $a$ and $t$ using a transparent model $f_\theta$. This approach ensures that the resulting model directly models treatment effects in an interpretable way, facilitating expert scrutiny and alignment with domain knowledge.

\subsection{Constructing the Surrogate Treatment Effects}
Given the definition $\tau_t(a)  = \mathbb{E}[Y_t(a) - Y_t(0) \vert A = a, T = t]$, our goal is to construct surrogate treatment effects $\tilde{\tau}_t(a) \approx Y_t(a) - Y_t(0)$, such that $\mathbb{E}[\tilde{\tau}_t(a) \vert A = a, T = t] = \tau_t(a)$. This approach mimics the concept of \textit{pseudo outcomes} used in CATE meta-learners \citep{kunzel_metalearners_2019, kennedy_towards_2023}, but poses additional challenges in the continuous dose-time setting.

\paragraph{For which pairs $(a, t)$ should we generate the surrogate treatment effect?} In the binary treatment case ($a \in \{0, 1\}$), each patient yields a single pseudo-outcome. However, with $\tau_t(a)$, each patient $i$ could have an infinite number of potential surrogate treatment effects $\tilde{\tau}_t(a)$ across all dose-time combinations. Generating excessive surrogate pairs risks over-reliance on the model used to estimate $\tilde{\tau}_t(a)$, which can introduce bias and obscure the information encoded in the original dataset $\mathcal{D}$. This, in turn, complicates the diagnosis of inconsistencies in the final estimate $\hat{\tau}_t(a)$. To maximise reliance on the original dataset and minimise the dependence on the potentially black-box model for $\tilde{\tau}_t(a)$, we propose to restrict the surrogate treatment effect generation to observed treatment-dose pairs: $\mathcal{P} := \{(A_i, T_{i, j}) \vert i \in [n], j \in [m_i], A_i > 0\}$. We exclude cases when $A_i = 0$ since by assumption $\tilde{\tau}_t(0) = 0$.

\paragraph{How to construct the surrogate treatment effect $\tilde{\tau}_t(a)$?} For each $(a, t) \in \mathcal{P}$, the observed outcome $Y_{i, j}$ is available. To construct the surrogate treatment effects $\tilde{\tau}_t(a)$, we thus only need to estimate the missing baseline outcome $Y_{T_{i, j}}(0)$. To maximise predictive accuracy, we propose to utilise the previously neglected patient covariates, $\bm{X}$. Specifically, for each sample $(\bm{X}_i, A_i, (T_{i, j}, Y_{i,j})_{j=1}^{m_i})$ with $A_i > 0$, we define $\tilde{\tau}_{i, j} = Y_{i, j} - \hat{\varphi}_0(\bm{X}_i, T_{i, j})$. Here, $\hat{\varphi}^0(\bm{X}_i, T_{i, j})$ is a \textit{baseline trajectory model}, trained using only the data of untreated patients: $\{i \in [n]: A_i = 0\}$.

In \cref{appendix: a0_model}, we empirically evaluate three alternative baseline trajectory models $\hat{\varphi}_0(\bm{X}_i, T_{i, j})$: two based on standard regression of the baseline outcome $Y_t(0)$ on the covariates $\bm{X}$ and the time $\bm{T}$, and one based on nearest neighbour matching in the $\mathcal{X}$-space.

\section{Interpretable Estimation of the Treatment Effects Using SemanticODE}
\label{sec: semantic}

As discussed in \cref{sec: dalte_transparent}, $\tau_t(a)$ supports transparent solutions because the key properties of each treatment effect trajectory -- such as the coordinates of its maximum and the value of its horizontal asymptote -- can be captured as univariate functions of the dose. We refer to these critical attributes as a \textit{semantic representation} of the trajectory.

\textit{Direct semantic modelling}, introduced by \citep{kacprzyk_no_2024}, aims to learn this semantic representation of a trajectory $x \in C^2(\mathcal{T})$ directly from data, which facilitates intuitive visualisation and subsequent analysis, straightforward expert edits, and the ability to impose domain-specific inductive biases. These features are particularly valuable for the estimation of $\tau_t(a)$, as transparency and domain alignment are crucial for adoption in high-stakes domains, such as clinical trials \citep{goodman_european_2017}.

\paragraph{SemanticODE Overview.} A key instantiation of the direct semantic modelling is the \textit{SemanticODE} method, which describes a trajectory, $x$, using two main components: the \textit{composition map} and the \textit{property  map}. Then the semantic representation of the trajectory predicted by those models is passed to a \textit{trajectory predictor} that predicts the trajectory based on its semantic representation. \\
\textbf{Composition Map.}
This map describes the overall \emph{shape} of the trajectory as a function of an initial condition $x(0) = x_0$. Specifically, it characterises each trajectory as a sequence of predefined ``motifs''. Each motif has the form $s_{\pm\pm*}$, i.e., is described by two symbols (each $+$ or $-$) and a letter (one of $b,u,h$). The two symbols describe the monotonicity and convexity of the trajectory segment. The letter $b$ signifies that the segment is bounded (between two \textit{transition points}), whereas $u$ and $h$ denote ``unbounded'' motifs describing a trajectory segment on some interval $(t,+\infty)$. See Figure 3 in \citet{kacprzyk_no_2024} for an illustration of the different motif types. For instance, $s_{+-h}$ describes a trajectory segment that is increasing ($+$), concave ($-$), and approaches a horizontal asymptote as $t\to\infty$ ($h$). The transition points between motifs correspond to local minima, maxima, and inflection points. The composition map divides the domain of $x_0$ into a few intervals and learns a different composition for each of them. For instance, a trajectory defined as $x(t) = x_0 e^t$ would have a composition map that assigns $(s_{++u})$ to $x_0 \geq 0$ and $(s_{--u})$ for $x_0 < 0$. \\
\textbf{Property Map.} Once the general shape is established, SemanticODE learns a \emph{property map} specifying how key properties of the identified composition depend on $x_0$. These properties include the coordinates of stationary and inflection points, the value of derivatives at boundary points, and additional characteristics of the last motif (e.g., the value of the horizontal asymptote).

\subsection{Adjusting SemanticODE to treatment effect estimation}

To make SemanticODE suitable for the estimation of $\tau_t(a)$, we introduce the following two adaptations. We term the resulting adaptation of SemanticODE as SemanticATE.

\circled{1} As the treatment effect trajectory of interest is never directly observed in the data, rather than fitting SemanticODE on the observed outcomes, $\{(T_{i, j}, Y_{i, j})_{j=1}^{m_i}\}_{i=1}^n$, we fit it to the surrogate treatment effects: $\{(T_{i, j}, \tilde{\tau}_{i, j})_{j=1}^{m_i}\}_{i=1}^n$. As described in the previous section, these surrogate effects approximate $Y_t(a) - Y_t(0)$ for each patient, and thus directly encode treatment effect trajectories. \\
\circled{2} In the original SemanticODE framework, the composition and property maps depend on the initial value $x_0$. Here, we instead use the dose $a$ as the controlling variable, defining both the composition maps and the property maps as a function $a$. This is necessary particularly as by definition, $\tau_0(a) = 0 \forall a \in \mathcal{A}$ (i.e. the treatment effect is zero at the time of treatment administration) invalidating the usual ``initial-condition'' approach commonly seen in ODE discovery methods \citep{brunton_discovering_2016}.
\section{Numerical Verification}
\label{sec: experiments}

\begin{figure*}
    \centering
    \includegraphics[width=\linewidth]{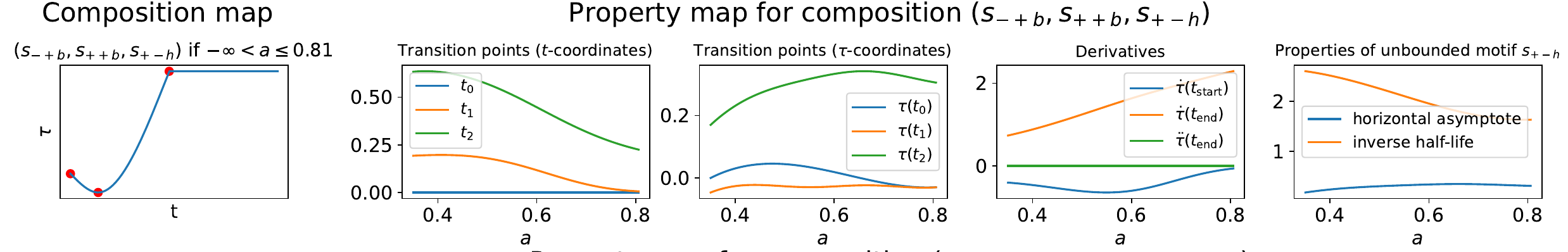}
    \vspace{-2em}
    \caption{A visualisation of a composition map and its corresponding property maps learned by SemanticATE.}
    \label{fig: composition_map}
\end{figure*}

The details of the experiments are provided in \cref{appendix: experiments}. Anonymised research code to reproduce the experiments can be found at: \url{https://anonymous.4open.science/r/Semantic-DALTE-4B52}.

\subsection{Details of the Experimental Setup}
The inability to observe counterfactual outcomes poses a big challenge for evaluation of causal models. While well-established synthetic benchmarks exist in binary treatment or static-continuous treatment settings, to the best of our knowledge there is no equivalent for continuous-dose continuous-time setting. To address this gap, we introduce three datasets that generate plausible dose–response trajectories, providing a controlled environment for the evaluation of $\tau_t(a)$ estimation methods.

\paragraph{Datasets.} \textit{PK Dataset.} We base this dataset on the pharmacokinetic (PK) model of tacrolimus plasma concentration proposed by \citet{woillard_population_2011}, which uses a system of ODEs to describe how an outcome evolves over time in response to a continuous dose of treatment, as well as several baseline covariates measured at $t=0$. We present the exact system of ODEs used to model the outcome in \cref{appendix: datasets}. We obtain two versions of this dataset: in \textit{PK-random}, we randomly sample the static patient covariates from a pre-specified range, while in \textit{PK-real} we use covariates from the dataset used in the original study. While $\tau_t(a)$ does not arise explicitly from the PK model, we approximate it by averaging counterfactual outcomes simulated for each patient: $\tau_t(a) = \frac{1}{n}\sum_{i=1}^{n}(y_t(a)_i - y_t(0)_i)$, where $y_t(a)_i$ is the true potential outcome for patient $i$ under dose $a$. \\
\textit{IHDP-based Dataset.} We construct this dataset using the same baseline covariates as the standard IHDP dataset \citep{hill_bayesian_2011}, obtaining the potential outcomes using:
$$y(t, a, \bm{x}) = \tau_t(a) + h(a, \bm{x}) + y^0_t(\bm{x}) + \epsilon.$$
Here, $h(a, \bm{x})$ is a \textit{static} heterogenous treatment effect function such that $\forall a \in \mathcal{A}$, $\mathbb{E}[h(a, \bm{X})] = 0$, $y^0_t(\bm{x})$ is a baseline outcome trajectory and $\epsilon$ is random noise. For $h(a, \bm{x})$ we use the continuous-treatment version of the IHDP dataset introduced by \cite{nie_vcnet_2021}. For $y^0_t(\bm{x})$ we use B-spline, the coefficients of which are obtained by multiplying the covariates $\bm{x}$ by randomly generated weights. Further details of both datasets, including parameter settings and generation procedures, are provided in \cref{appendix: datasets}.

\paragraph{Metrics.} Following the evaluation practices for ADRF \citep{schwab_learning_2020}, we propose to evaluate the $\tau_t(a)$ estimation accuracy using the \textbf{mean integrated square error}, which we define as:
$$\textrm{MISE} = \int_0^{t_{max}} \int_{a_{min}}^{a_{max}}(\tau_t(a) - \hat{\tau}_t(a))^2 da \, dt.$$
We approximate this metric using:
$$\widehat{\textrm{MISE}} = \frac{1}{n_t \cdot n_a} \sum_{i = 1}^{n_t} \sum_{j = 1}^{n_a} (\tau_{t_i}(a_j) - \hat{\tau}_{t_i}(a_j))^2,$$
where $(t_1=0, t_2, \dots, t_{n_t}=t_{max})$ is a grid of $n_t$ equally distributed time points and $(a_1=a_{min}, a_2, \dots, a_{n_a}=a_{max})$ is a grid of $n_a$ equally distributed doses.
For each dataset, we standardise the values of $\tau_t(a)$ to have the standard deviation of $1.0$ before reporting the $\widehat{\textrm{MISE}}$ values.

\paragraph{Baseline trajectory models.} We use the XGBoost model, trained to predict the value $Y_t(0)$ from the concatenated variables $(\bm{x}, t)$, as our default baseline trajectory estimator. We provide a sensitivity analysis of how the choice of the baseline trajectory model impacts $\tau_t(a)$ estimation performance is presented in in the \cref{appendix: a0_model}.

\begin{table*}[ht!]
\centering
\small
\resizebox{\linewidth}{!}{
\begin{tabular}{lcccccc}
\toprule
Method & \multicolumn{2}{c}{PK-random} & \multicolumn{2}{c}{PK-real} & \multicolumn{2}{c}{IHDP-based} \\
 & In-domain & Out-domain & In-domain & Out-domain & In-domain & Out-domain \\
\midrule
SINDy & 2.81 ± 1.04 & 3.88 ± 5.62 & $>10^6$ & $>10^3$ & 0.46 ± 0.12 & 2.32 ± 0.54 \\
WSINDy & 0.53 ± 0.21 & 7.96 ± 3.64 & 1.31 ± 0.50 & 5.21 ± 1.27 & 0.54 ± 0.04 & 25.16 ± 31.20 \\
NeuralODE & 0.50 ± 0.03 & 0.17 ± 0.08 & 0.54 ± 0.12 & 0.21 ± 0.13 & 0.39 ± 0.19 & 1.37 ± 1.15 \\
XGBoost & 0.40 ± 0.15 & 0.11 ± 0.04 & 0.43 ± 0.35 & 0.08 ± 0.02 & 0.36 ± 0.14 & 0.20 ± 0.09 \\
PolyReg & \textbf{0.27 ± 0.06} & 211.05 ± 82.49 & \textbf{0.34 ± 0.10} & 197.25 ± 69.20 & 0.05 ± 0.01 & 4.31 ± 1.83 \\
\hline
SemanticATE & 0.41 ± 0.28 & \textbf{0.04 ± 0.01} & 0.37 ± 0.29 & \textbf{0.02 ± 0.01} & \textbf{0.04 ± 0.02} & \textbf{0.02 ± 0.02} \\

\bottomrule
\end{tabular}}
\caption{Comparison of different $\tau_t(a)$ estimation strategies. We report the average in-domain ($t \in [0, 1]$) and out-domain ($t \in [1, 1.25]$) $\widehat{MISE} \pm$ std, computed over 5 seeds.}
\label{tab: trajectory_models}
\end{table*}

\subsection{Transparent estimation with SemanticATE}
Here, we showcase SemanticATE's ability to generate transparent and verifiable predictions of $\tau_t(a)$. 

\paragraph{Verification and Editing.} We note that the naive implementation of SemanticATE can be fitted without any constraints, allowing to find the composition and property maps which best fit the provided dataset. However, further modifications can also be introduced to align the model with domain knowledge: \\
\circled{1} \textit{Inductive Biases (IB).} The practitioner can easily impose the necessary inductive biases by narrowing down the set of possible compositions (i.e. constraining the shape of the treatment effect trajectories). For our datasets, we enforce that the each composition ends in a horizontal asymptote, so the treatment effect decays to some fixed value over time (rather than exploding towards infinity). \\
\circled{2} \textit{Editing.} SemanticATE also allows the practitioner to scrutinise the model post-training and introduce edits if the results do not agree with domain knowledge. In \cref{fig: composition_map} we showcase one of the composition maps, and its corresponding property maps, obtained by fitting the constrained SemanticATE to the \textit{IHDP-based} dataset. We note that in the fitted model, $\tau_0(a)$, visualised in the second property map graph as $\uptau(t_0)$, is different from zero. As this contradicts our assumption that the treatment effect should be zero at the time of treatment administration $t=0$, we can manually modify this property map.

As we show in \cref{table: semantic_dalte_demo}, introducing these constraints and modifications which ensure alignment with domain-knowledge does not significantly deteriorate the performance of the model.

\begin{table}[h]
\vspace{-1em}
\centering
\small
\begin{tabular}{lccc}
\toprule
Method & PK-random & PK-real & IHDP-based \\
\midrule
Base & 0.25 ± 0.09 & 0.22 ± 0.10 & 0.04 ± 0.02 \\
Base + IB & 0.25 ± 0.10 & 0.23 ± 0.09 & 0.06 ± 0.01 \\
Base + IB + edits & 0.23 ± 0.08 & 0.23 ± 0.09 & 0.08 ± 0.03 \\
\bottomrule
\end{tabular}
\caption{Performance of the SemanticATE model with and without modifications. We report the mean $\widehat{MISE}$ $\pm$ \textit{std}, computed over 10 seeds.}
\label{table: semantic_dalte_demo}
\vspace{-1.5em}
\end{table}

\paragraph{Intuitive analysis.} Modelling $\tau_t(a)$ through the composition and property maps facilitates not only verification and editing, but also  easy analysis of the resulting model. Looking at \cref{fig: composition_map} we see that, as expected, the value of the maximum treatment effect increases as we increase the dosage, with a plateau reached around $a=0.6$, while the time to reach the maximum decreases. We can also see that the value of the horizontal asymptote plateaus around $a=0.6$. These insights might help the practitioners choose the optimal dosage, avoiding over-treatment.

\begin{figure}[h]
    \centering
    \includegraphics[width=0.8\linewidth]{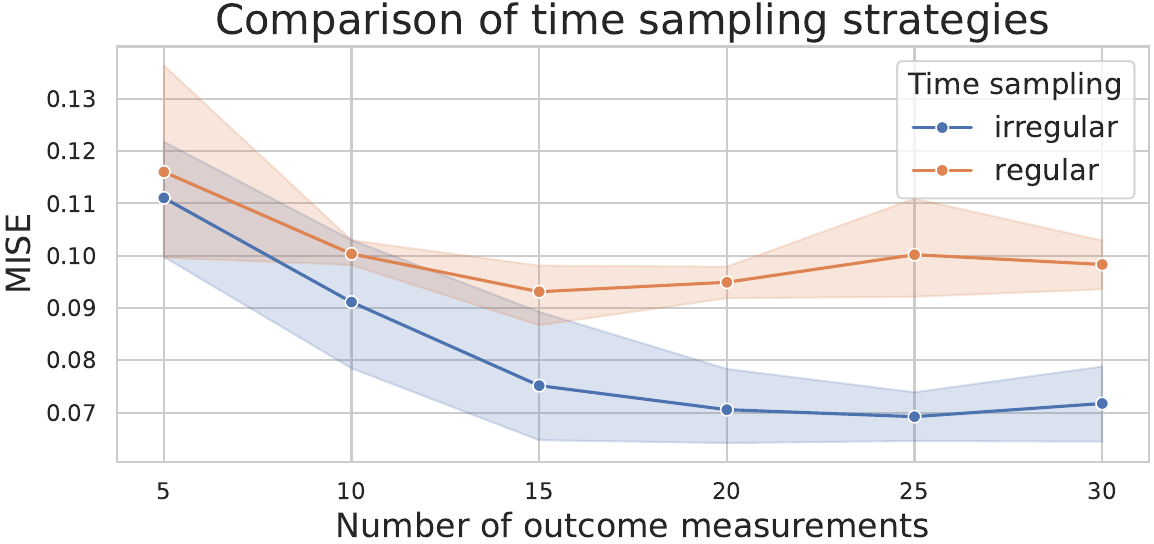}
    \caption{Performance of SemanticATE with the changing number of outcome measurements. }
    \label{fig: time_sampling}
\end{figure}

\subsection{Comparison to alternative estimation methods}
We demonstrate that, apart from being transparent-by-design, the benefit of using SemanticODE for the $\tau_t(a)$ estimation task lies in its good performance. Here, we compare to alternative methods for modelling $\tau_t(a)$. We consider standard ODE discovery algorithms, SINDy and WSINDy, as implemented by the \texttt{PySINDy} library, and three black-box approaches: NeuralODE, XGBoost and polynomial regression. To make the ODE discovery methods suitable for $\tau_t(a)$ estimation setting, we fix the timing of the first observation $t_0$ across all individuals, and fit an additional model to predict $\tau_{t_0}(a)$ based on the dose $a$. $\tau_{t_0}(a)$ is then used as an initial condition for the trajectory. Because of the requirements of the NeuralODE method, for this experiment we rely on \textit{regular} sampling of the outcomes.

\textbf{Results.} SemanticATE consistently outperforms the glass-box methods (SINDy and WSINDy). While in some cases the black box models outperform SemanticATE \textit{in-domain}, by imposing the necessary inductive biases (in this case, we used the fact that the trajectory ends in a horizontal asymptote), SemanticATE achieves incomparable out-domain performance.

\subsection{Insight: Efficient Trial Design}
Although $\tau_t(a)$ can be broadly applied, it holds particular relevance and potential for RCTs. In early-phase clinical trials -- where patient safety is paramount and data is scarce -- precisely determining the optimal dosage is both crucial and challenging. As methods employed in clinical trials need to be characterised by sample efficiency, we analyse the performance of SemanticATE as we change the number of patients (see \cref{appendix: n_samples}) and as we change the number of outcome measurements per patient, presented below.

\textbf{Results.} The results, presented in Figure \cref{fig: time_sampling}, show that SemanticATE reaches optimal performance with only 15 outcome measurements. Further, we note that the performance of the model improves significantly when irregular (random) sampling of the measurement times is employed. This insight can guide the design of the clinical trials.
\section{Discussion and Future Work}
\label{sec: discussion}
Modelling the treatment effect as a smooth function of dose and time through the use of SemanticODE reorients time-varying causal inference toward interpretability and robustness, positioning it for applications in high-stakes domains. While in this preliminary paper we have assumed that the data used for the estimation of $\tau_t(a)$ comes from an RCT, a natural direction for future work would be to consider estimation from \textbf{observational data}, where non-random assignment of the treatment dose or the measurement times can lead to bias when not accounted for. Further, we believe that introducing a notion of \textbf{uncertainty quantification} into the predictions obtained from SemanticATE, for example by quantifying the learned uncertainty of the property maps, could further allow to make SemanticATE more suitable for high-stake domains.

\bibliography{references}

\begin{thebibliography}{36}
\providecommand{\natexlab}[1]{#1}
\providecommand{\url}[1]{\texttt{#1}}
\expandafter\ifx\csname urlstyle\endcsname\relax
  \providecommand{\doi}[1]{doi: #1}\else
  \providecommand{\doi}{doi: \begingroup \urlstyle{rm}\Url}\fi

\bibitem[Bica et~al.(2019)Bica, Alaa, Jordon, and Schaar]{bica_estimating_2019}
Bica, I., Alaa, A.~M., Jordon, J., and Schaar, M. v.~d.
\newblock Estimating counterfactual treatment outcomes over time through adversarially balanced representations.
\newblock September 2019.
\newblock URL \url{https://openreview.net/forum?id=BJg866NFvB}.

\bibitem[Brunton et~al.(2016)Brunton, Proctor, and Kutz]{brunton_discovering_2016}
Brunton, S.~L., Proctor, J.~L., and Kutz, J.~N.
\newblock Discovering governing equations from data by sparse identification of nonlinear dynamical systems.
\newblock \emph{Proceedings of the National Academy of Sciences}, 113\penalty0 (15):\penalty0 3932--3937, April 2016.
\newblock \doi{10.1073/pnas.1517384113}.
\newblock URL \url{https://www.pnas.org/doi/10.1073/pnas.1517384113}.
\newblock Publisher: Proceedings of the National Academy of Sciences.

\bibitem[Chen et~al.(2018)Chen, Rubanova, Bettencourt, and Duvenaud]{chen_neural_2018}
Chen, R. T.~Q., Rubanova, Y., Bettencourt, J., and Duvenaud, D.~K.
\newblock Neural {Ordinary} {Differential} {Equations}.
\newblock In \emph{Advances in {Neural} {Information} {Processing} {Systems}}, volume~31. Curran Associates, Inc., 2018.
\newblock URL \url{https://proceedings.neurips.cc/paper/2018/hash/69386f6bb1dfed68692a24c8686939b9-Abstract.html}.

\bibitem[Chen \& Guestrin(2016)Chen and Guestrin]{chen_xgboost_2016}
Chen, T. and Guestrin, C.
\newblock {XGBoost}: {A} {Scalable} {Tree} {Boosting} {System}.
\newblock In \emph{Proceedings of the 22nd {ACM} {SIGKDD} {International} {Conference} on {Knowledge} {Discovery} and {Data} {Mining}}, pp.\  785--794, August 2016.
\newblock \doi{10.1145/2939672.2939785}.
\newblock URL \url{http://arxiv.org/abs/1603.02754}.
\newblock arXiv:1603.02754 [cs].

\bibitem[Curth \& van~der Schaar(2021)Curth and van~der Schaar]{curth_inductive_2021}
Curth, A. and van~der Schaar, M.
\newblock On {Inductive} {Biases} for {Heterogeneous} {Treatment} {Effect} {Estimation}.
\newblock In \emph{Advances in {Neural} {Information} {Processing} {Systems}}, volume~34, pp.\  15883--15894. Curran Associates, Inc., 2021.
\newblock URL \url{https://proceedings.neurips.cc/paper/2021/hash/8526e0962a844e4a2f158d831d5fddf7-Abstract.html}.

\bibitem[Goodman \& Flaxman(2017)Goodman and Flaxman]{goodman_european_2017}
Goodman, B. and Flaxman, S.
\newblock European {Union} {Regulations} on {Algorithmic} {Decision}-{Making} and a “{Right} to {Explanation}”.
\newblock \emph{AI Magazine}, 38\penalty0 (3):\penalty0 50--57, October 2017.
\newblock ISSN 2371-9621.
\newblock \doi{10.1609/aimag.v38i3.2741}.
\newblock URL \url{https://ojs.aaai.org/aimagazine/index.php/aimagazine/article/view/2741}.
\newblock Number: 3.

\bibitem[Han et~al.(2018)Han, Lee, and Pang]{han_finding_2018}
Han, Y.~R., Lee, P.~I., and Pang, K.~S.
\newblock Finding {Tmax} and {Cmax} in {Multicompartmental} {Models}.
\newblock \emph{Drug Metabolism and Disposition: The Biological Fate of Chemicals}, 46\penalty0 (11):\penalty0 1796--1804, November 2018.
\newblock ISSN 1521-009X.
\newblock \doi{10.1124/dmd.118.082636}.

\bibitem[Hess et~al.(2024)Hess, Melnychuk, Frauen, and Feuerriegel]{hess_bayesian_2024}
Hess, K., Melnychuk, V., Frauen, D., and Feuerriegel, S.
\newblock Bayesian {Neural} {Controlled} {Differential} {Equations} for {Treatment} {Effect} {Estimation}, April 2024.
\newblock URL \url{http://arxiv.org/abs/2310.17463}.
\newblock arXiv:2310.17463.

\bibitem[Hill(2011)]{hill_bayesian_2011}
Hill, J.~L.
\newblock Bayesian {Nonparametric} {Modeling} for {Causal} {Inference}.
\newblock \emph{Journal of Computational and Graphical Statistics}, 20\penalty0 (1):\penalty0 217--240, March 2011.
\newblock ISSN 10618600.
\newblock \doi{10.1198/JCGS.2010.08162}.
\newblock URL \url{https://www.tandfonline.com/doi/abs/10.1198/jcgs.2010.08162}.
\newblock Publisher: Taylor \& Francis.

\bibitem[Hirano \& Imbens(2004)Hirano and Imbens]{hirano_propensity_2004}
Hirano, K. and Imbens, G.~W.
\newblock The {Propensity} {Score} with {Continuous} {Treatments}.
\newblock In \emph{Applied {Bayesian} {Modeling} and {Causal} {Inference} from {Incomplete}-{Data} {Perspectives}}, pp.\  73--84. John Wiley \& Sons, Ltd, 2004.
\newblock ISBN 978-0-470-09045-9.
\newblock \doi{10.1002/0470090456.ch7}.
\newblock URL \url{https://onlinelibrary.wiley.com/doi/abs/10.1002/0470090456.ch7}.
\newblock Section: 7 \_eprint: https://onlinelibrary.wiley.com/doi/pdf/10.1002/0470090456.ch7.

\bibitem[Kacprzyk \& Van Der~Schaar(2024)Kacprzyk and Van Der~Schaar]{kacprzyk_no_2024}
Kacprzyk, K. and Van Der~Schaar, M.
\newblock No {Equations} {Needed}: {Learning} {System} {Dynamics} {Without} {Relying} on {Closed}-{Form} {ODEs}.
\newblock October 2024.
\newblock URL \url{https://arxiv.org/abs/2501.18563}.

\bibitem[Kacprzyk et~al.(2024)Kacprzyk, Holt, Berrevoets, Qian, and van~der Schaar]{kacprzyk_ode_2024}
Kacprzyk, K., Holt, S., Berrevoets, J., Qian, Z., and van~der Schaar, M.
\newblock {ODE} {Discovery} for {Longitudinal} {Heterogeneous} {Treatment} {Effects} {Inference}, March 2024.
\newblock URL \url{http://arxiv.org/abs/2403.10766}.
\newblock arXiv:2403.10766 [cs, stat].

\bibitem[Kaptanoglu et~al.(2022)Kaptanoglu, Silva, Fasel, Kaheman, Goldschmidt, Callaham, Delahunt, Nicolaou, Champion, Loiseau, Kutz, and Brunton]{kaptanoglu_pysindy_2022}
Kaptanoglu, A.~A., Silva, B. M.~d., Fasel, U., Kaheman, K., Goldschmidt, A.~J., Callaham, J., Delahunt, C.~B., Nicolaou, Z.~G., Champion, K., Loiseau, J.-C., Kutz, J.~N., and Brunton, S.~L.
\newblock {PySINDy}: {A} comprehensive {Python} package for robust sparse system identification.
\newblock \emph{Journal of Open Source Software}, 7\penalty0 (69):\penalty0 3994, January 2022.
\newblock ISSN 2475-9066.
\newblock \doi{10.21105/joss.03994}.
\newblock URL \url{https://joss.theoj.org/papers/10.21105/joss.03994}.

\bibitem[Kennedy(2023)]{kennedy_towards_2023}
Kennedy, E.~H.
\newblock Towards optimal doubly robust estimation of heterogeneous causal effects.
\newblock \emph{Electronic Journal of Statistics}, 17\penalty0 (2):\penalty0 3008--3049, January 2023.
\newblock ISSN 1935-7524, 1935-7524.
\newblock \doi{10.1214/23-EJS2157}.
\newblock URL \url{https://projecteuclid.org/journals/electronic-journal-of-statistics/volume-17/issue-2/Towards-optimal-doubly-robust-estimation-of-heterogeneous-causal-effects/10.1214/23-EJS2157.full}.
\newblock Publisher: Institute of Mathematical Statistics and Bernoulli Society.

\bibitem[Künzel et~al.(2019)Künzel, Sekhon, Bickel, and Yu]{kunzel_metalearners_2019}
Künzel, S.~R., Sekhon, J.~S., Bickel, P.~J., and Yu, B.
\newblock Metalearners for estimating heterogeneous treatment effects using machine learning.
\newblock \emph{Proceedings of the National Academy of Sciences}, 116\penalty0 (10):\penalty0 4156--4165, March 2019.
\newblock \doi{10.1073/pnas.1804597116}.
\newblock URL \url{https://www.pnas.org/doi/abs/10.1073/pnas.1804597116}.
\newblock Publisher: Proceedings of the National Academy of Sciences.

\bibitem[Li et~al.(2021)Li, Hu, Lu, Utsumi, Chakraborty, Sow, Madan, Li, Ghalwash, Shahn, and Lehman]{li_g-net_2021}
Li, R., Hu, S., Lu, M., Utsumi, Y., Chakraborty, P., Sow, D.~M., Madan, P., Li, J., Ghalwash, M., Shahn, Z., and Lehman, L.-w.
\newblock G-{Net}: a {Recurrent} {Network} {Approach} to {G}-{Computation} for {Counterfactual} {Prediction} {Under} a {Dynamic} {Treatment} {Regime}.
\newblock In \emph{Proceedings of {Machine} {Learning} for {Health}}, pp.\  282--299. PMLR, November 2021.
\newblock URL \url{https://proceedings.mlr.press/v158/li21a.html}.
\newblock ISSN: 2640-3498.

\bibitem[Liao et~al.(2020)Liao, Gruen, and Miller]{liao_questioning_2020}
Liao, Q.~V., Gruen, D., and Miller, S.
\newblock Questioning the {AI}: {Informing} {Design} {Practices} for {Explainable} {AI} {User} {Experiences}.
\newblock In \emph{Proceedings of the 2020 {CHI} {Conference} on {Human} {Factors} in {Computing} {Systems}}, pp.\  1--15, April 2020.
\newblock \doi{10.1145/3313831.3376590}.
\newblock URL \url{http://arxiv.org/abs/2001.02478}.
\newblock arXiv:2001.02478 [cs].

\bibitem[Lin et~al.(2004)Lin, Scharfstein, and Rosenheck]{lin_analysis_2004}
Lin, H., Scharfstein, D.~O., and Rosenheck, R.~A.
\newblock Analysis of {Longitudinal} {Data} with {Irregular}, {Outcome}-{Dependent} {Follow}-{Up}.
\newblock \emph{Journal of the Royal Statistical Society Series B: Statistical Methodology}, 66\penalty0 (3):\penalty0 791--813, August 2004.
\newblock ISSN 1369-7412.
\newblock \doi{10.1111/j.1467-9868.2004.b5543.x}.
\newblock URL \url{https://doi.org/10.1111/j.1467-9868.2004.b5543.x}.

\bibitem[Melnychuk et~al.(2022)Melnychuk, Frauen, and Feuerriegel]{melnychuk_causal_2022}
Melnychuk, V., Frauen, D., and Feuerriegel, S.
\newblock Causal {Transformer} for {Estimating} {Counterfactual} {Outcomes}.
\newblock In \emph{Proceedings of the 39th {International} {Conference} on {Machine} {Learning}}, pp.\  15293--15329. PMLR, June 2022.
\newblock URL \url{https://proceedings.mlr.press/v162/melnychuk22a.html}.
\newblock ISSN: 2640-3498.

\bibitem[Messenger \& Bortz(2021)Messenger and Bortz]{messenger_weak_2021}
Messenger, D.~A. and Bortz, D.~M.
\newblock Weak {SINDy}: {Galerkin}-{Based} {Data}-{Driven} {Model} {Selection}.
\newblock \emph{Multiscale Modeling \& Simulation}, 19\penalty0 (3):\penalty0 1474--1497, January 2021.
\newblock ISSN 1540-3459, 1540-3467.
\newblock \doi{10.1137/20M1343166}.
\newblock URL \url{http://arxiv.org/abs/2005.04339}.
\newblock arXiv:2005.04339 [math].

\bibitem[Moodie \& Stephens(2012)Moodie and Stephens]{moodie_estimation_2012}
Moodie, E. E.~M. and Stephens, D.~A.
\newblock Estimation of dose-response functions for longitudinal data using the generalised propensity score.
\newblock \emph{Statistical Methods in Medical Research}, 21\penalty0 (2):\penalty0 149--166, April 2012.
\newblock ISSN 1477-0334.
\newblock \doi{10.1177/0962280209340213}.

\bibitem[Nagalapatti et~al.(2024)Nagalapatti, Iyer, De, and Sarawagi]{nagalapatti_continuous_2024}
Nagalapatti, L., Iyer, A., De, A., and Sarawagi, S.
\newblock Continuous {Treatment} {Effect} {Estimation} {Using} {Gradient} {Interpolation} and {Kernel} {Smoothing}, January 2024.
\newblock URL \url{http://arxiv.org/abs/2401.15447}.
\newblock arXiv:2401.15447 [cs, stat].

\bibitem[Nie et~al.(2021)Nie, Ye, Liu, and Nicolae]{nie_vcnet_2021}
Nie, L., Ye, M., Liu, Q., and Nicolae, D.
\newblock {VCNet} and {Functional} {Targeted} {Regularization} {For} {Learning} {Causal} {Effects} of {Continuous} {Treatments}, March 2021.
\newblock URL \url{http://arxiv.org/abs/2103.07861}.
\newblock arXiv:2103.07861 [cs].

\bibitem[Qian et~al.(2021)Qian, Zhang, Bica, Wood, and van~der Schaar]{qian_synctwin_2021}
Qian, Z., Zhang, Y., Bica, I., Wood, A., and van~der Schaar, M.
\newblock {SyncTwin}: {Treatment} {Effect} {Estimation} with {Longitudinal} {Outcomes}.
\newblock In \emph{Advances in {Neural} {Information} {Processing} {Systems}}, volume~34, pp.\  3178--3190. Curran Associates, Inc., 2021.
\newblock URL \url{https://proceedings.neurips.cc/paper_files/paper/2021/hash/19485224d128528da1602ca47383f078-Abstract.html}.

\bibitem[Reinbold et~al.(2020)Reinbold, Gurevich, and Grigoriev]{reinbold_using_2020}
Reinbold, P. A.~K., Gurevich, D.~R., and Grigoriev, R.~O.
\newblock Using {Noisy} or {Incomplete} {Data} to {Discover} {Models} of {Spatiotemporal} {Dynamics}.
\newblock \emph{Physical Review E}, 101\penalty0 (1):\penalty0 010203, January 2020.
\newblock ISSN 2470-0045, 2470-0053.
\newblock \doi{10.1103/PhysRevE.101.010203}.
\newblock URL \url{http://arxiv.org/abs/1911.03365}.
\newblock arXiv:1911.03365 [math].

\bibitem[Robins(1986)]{robins_new_1986}
Robins, J.
\newblock A new approach to causal inference in mortality studies with a sustained exposure period—application to control of the healthy worker survivor effect.
\newblock \emph{Mathematical Modelling}, 7\penalty0 (9):\penalty0 1393--1512, January 1986.
\newblock ISSN 0270-0255.
\newblock \doi{10.1016/0270-0255(86)90088-6}.
\newblock URL \url{https://www.sciencedirect.com/science/article/pii/0270025586900886}.

\bibitem[Robins(1994)]{robins_correcting_1994}
Robins, J.~M.
\newblock Correcting for non-compliance in randomized trials using structural nested mean models.
\newblock \emph{Communications in Statistics - Theory and Methods}, 23\penalty0 (8):\penalty0 2379--2412, January 1994.
\newblock ISSN 0361-0926.
\newblock \doi{10.1080/03610929408831393}.
\newblock URL \url{https://doi.org/10.1080/03610929408831393}.
\newblock Publisher: Taylor \& Francis \_eprint: https://doi.org/10.1080/03610929408831393.

\bibitem[Robins et~al.(1995)Robins, Rotnitzky, and Zhao]{robins_analysis_1995}
Robins, J.~M., Rotnitzky, A., and Zhao, L.~P.
\newblock Analysis of {Semiparametric} {Regression} {Models} for {Repeated} {Outcomes} in the {Presence} of {Missing} {Data}.
\newblock \emph{Journal of the American Statistical Association}, 90\penalty0 (429):\penalty0 106--121, March 1995.
\newblock ISSN 0162-1459.
\newblock \doi{10.1080/01621459.1995.10476493}.
\newblock URL \url{https://www.tandfonline.com/doi/abs/10.1080/01621459.1995.10476493}.
\newblock Publisher: ASA Website \_eprint: https://www.tandfonline.com/doi/pdf/10.1080/01621459.1995.10476493.

\bibitem[Robins et~al.(2000)Robins, Hernán, and Brumback]{robins_marginal_2000}
Robins, J.~M., Hernán, M.~A., and Brumback, B.
\newblock Marginal structural models and causal inference in epidemiology.
\newblock \emph{Epidemiology (Cambridge, Mass.)}, 11\penalty0 (5):\penalty0 550--560, September 2000.
\newblock ISSN 1044-3983.
\newblock \doi{10.1097/00001648-200009000-00011}.

\bibitem[Rubin(1984)]{rubin_bayesianly_1984}
Rubin, D.~B.
\newblock Bayesianly {Justifiable} and {Relevant} {Frequency} {Calculations} for the {Applied} {Statistician}.
\newblock \emph{The Annals of Statistics}, 12\penalty0 (4):\penalty0 1151--1172, 1984.
\newblock ISSN 0090-5364.
\newblock URL \url{https://www.jstor.org/stable/2240995}.
\newblock Publisher: Institute of Mathematical Statistics.

\bibitem[Schwab et~al.(2020)Schwab, Linhardt, Bauer, Buhmann, and Karlen]{schwab_learning_2020}
Schwab, P., Linhardt, L., Bauer, S., Buhmann, J.~M., and Karlen, W.
\newblock Learning {Counterfactual} {Representations} for {Estimating} {Individual} {Dose}-{Response} {Curves}.
\newblock \emph{Proceedings of the AAAI Conference on Artificial Intelligence}, 34\penalty0 (04):\penalty0 5612--5619, April 2020.
\newblock ISSN 2374-3468.
\newblock \doi{10.1609/aaai.v34i04.6014}.
\newblock URL \url{https://ojs.aaai.org/index.php/AAAI/article/view/6014}.
\newblock Number: 04.

\bibitem[Seedat et~al.(2022)Seedat, Imrie, Bellot, Qian, and Schaar]{seedat_continuous-time_2022}
Seedat, N., Imrie, F., Bellot, A., Qian, Z., and Schaar, M. v.~d.
\newblock Continuous-{Time} {Modeling} of {Counterfactual} {Outcomes} {Using} {Neural} {Controlled} {Differential} {Equations}.
\newblock In \emph{Proceedings of the 39th {International} {Conference} on {Machine} {Learning}}, pp.\  19497--19521. PMLR, June 2022.
\newblock URL \url{https://proceedings.mlr.press/v162/seedat22b.html}.
\newblock ISSN: 2640-3498.

\bibitem[Silva et~al.(2020)Silva, Champion, Quade, Loiseau, Kutz, and Brunton]{silva_pysindy_2020}
Silva, B. M.~d., Champion, K., Quade, M., Loiseau, J.-C., Kutz, J.~N., and Brunton, S.~L.
\newblock {PySINDy}: {A} {Python} package for the sparse identification of nonlinear dynamical systems from data.
\newblock \emph{Journal of Open Source Software}, 5\penalty0 (49):\penalty0 2104, May 2020.
\newblock ISSN 2475-9066.
\newblock \doi{10.21105/joss.02104}.
\newblock URL \url{https://joss.theoj.org/papers/10.21105/joss.02104}.

\bibitem[Vanderschueren et~al.(2023)Vanderschueren, Curth, Verbeke, and Van Der~Schaar]{vanderschueren_accounting_2023}
Vanderschueren, T., Curth, A., Verbeke, W., and Van Der~Schaar, M.
\newblock Accounting for informative sampling when learning to forecast treatment outcomes over time.
\newblock In \emph{Proceedings of the 40th {International} {Conference} on {Machine} {Learning}}, volume 202 of \emph{{ICML}'23}, pp.\  34855--34874, Honolulu, Hawaii, USA, July 2023. JMLR.org.

\bibitem[Wang et~al.(2022)Wang, Lyu, Wu, Wu, and Chen]{wang_generalization_2022}
Wang, X., Lyu, S., Wu, X., Wu, T., and Chen, H.
\newblock Generalization {Bounds} for {Estimating} {Causal} {Effects} of {Continuous} {Treatments}.
\newblock \emph{Advances in Neural Information Processing Systems}, 35:\penalty0 8605--8617, December 2022.
\newblock URL \url{https://proceedings.neurips.cc/paper_files/paper/2022/hash/390bb66a088d37f62ee9fb779c5953c2-Abstract-Conference.html}.

\bibitem[Woillard et~al.(2011)Woillard, de~Winter, Kamar, Marquet, Rostaing, and Rousseau]{woillard_population_2011}
Woillard, J.-B., de~Winter, B. C.~M., Kamar, N., Marquet, P., Rostaing, L., and Rousseau, A.
\newblock Population pharmacokinetic model and {Bayesian} estimator for two tacrolimus formulations--twice daily {Prograf} and once daily {Advagraf}.
\newblock \emph{British Journal of Clinical Pharmacology}, 71\penalty0 (3):\penalty0 391--402, March 2011.
\newblock ISSN 1365-2125.
\newblock \doi{10.1111/j.1365-2125.2010.03837.x}.

\end{thebibliography}
\bibliographystyle{icml2025}

\newpage
\appendix
\onecolumn

\section{Extended Related Works}
\label{appendix: related}

\begin{table}[]
    \centering
    \begin{tabular}{cccccc}
    \hline
       \# & Examples & Level & Treatments & Outcomes & Transparent \\
       \hline
       1 & \citet{robins_marginal_2000, robins_new_1986, robins_correcting_1994} & Population & Sequential & Regular & \cmark \\
       2 & \citet{qian_synctwin_2021} & Personalised & Point & Regular & \xmark \\
       3 & \citet{bica_estimating_2019, li_g-net_2021, melnychuk_causal_2022} & Personalised & Sequential & Regular & \xmark \\
       4 & \citet{seedat_continuous-time_2022, hess_bayesian_2024} &  Personalised & Sequential & Irregular & \xmark \\
       5 & This work & Population & Point & Irregular & \cmark \\
        \hline
         
    \end{tabular}
    \caption{Table summarising the different settings considered in the time-varying treatment effect estimation literature.}
    \label{tab: related}
\end{table}

\circled{1} We summarise the different settings in the time-varying treatment effect estimation in \cref{tab: related}. We note that population-level causal estimands focusing on regularly sampled treatments and outcomes can be easily interpreted and verified if the number of time points is reasonable (because they provide $d^t$ alternative average potential outcomes, where $d$ is the number of possible categorical treatments, typically $d=2$). However, the insights they provide are limited, as they do not allow to clearly understand what happens \emph{between} the different interventions.

\circled{2} \citet{qian_synctwin_2021} has considered point treatments in the context of time-varying treatment effect estimation before, however their work focuses on estimating the conditional average potential outcome, and as such does not lend itself to transparent and verifiable analysis of the \emph{whole} model (it does not provides insights into the shape of the trajectory as a function of the available covariates).

\circled{3} This setting further extends the setting considered in \circled{1}, by aiming to provide personalised treatment effect predictions to individuals based on their covariate and treatment assignment history. However, being even more complex than the setting considered in \circled{2}, analysis, understanding and verification of models proposed in this setting is extremely difficult, if not impossible.

\circled{4} This setting further extends on the previous one, by allowing the treatment assignments, outcome and covariate measurements to be taken at different time points. By modelling the dynamics of the covariates, outcomes and treatments as continuous-time trajectories, these models aim to provide more insight into the evolution of the potential outcomes and treatment effects over time. However, yet again, by increasing the complexity of the setting (and allowing the treatments to happen at different time points for different individuals), this setting moves even further away from interpretability and transparency.

Finally, we note that while \citet{moodie_estimation_2012}, similarly as this work, consider the estimation of the average potential outcomes of a point continuous treatment from repeated outcome data, we note that they assume that the potential outcome functions do not change over time.

\section{Estimating $\tau_t(a)$ from Observational Data}
\label{appendix: estimation}

\subsection{What if the treatment assignment is non-random?}
In case when the treatment is assigned non-randomly, point identification of $\tau_t(a)$ from observational data is possible under the following two assumptions:
\begin{assumption}[Weak Unconfoundedness]
    For all times $t \in \mathcal{T}$ and for all dosage assignments $a \in \mathcal{A}$, the potential outcome $Y_a(t)$ is conditionally independent from the treatment given the covariates $\bm{X}$: $Y_t(a) \perp A \vert \bm{X}$.
\end{assumption}
Following \cite{hirano_propensity_2004, moodie_estimation_2012}, we refer to this as \emph{weak} unconfoundedness as we do not require joint independence of all potential outcomes, $\{Y_t(a): t \in \mathcal{T}, a \in \mathcal{A}\}$. We further need to make the overlap assumption.
\begin{assumption}[Treatment Overlap]
    The dosage assignment is non-deterministic: there exists a constant $c > 0$ such that $\forall \bm{x} \in \mathcal{X}$ such that $P(\bm{X} = \bm{x}) > 0$, we have $c < P(A = a \vert \bm{X}=\bm{x}) < 1 - c$.
\end{assumption}
Drawing inspiration from the literature on the continuous treatment effect estimation \cite{nie_vcnet_2021}, we note that under these two assumptions, at each time point $(a, t)$, $\tau_t(a)$ could be estimated as:
$$\tau_t(a) = \frac{1}{n}\sum_{i=1}^n (\mu(a, t, \bm{X}_i) - \mu(0, t, \bm{X}_i),$$
where $\mu(a, t, \bm{x}) = \mathbb{E}[Y \vert A = a, T =t, \bm{X} = \bm{x}]$. This estimation method would require us to obtain a model $\mu(a, t, \bm{x})$.
Even in the static time, when the goal is to obtain the model $\mu(a, \bm{x})$, this task is challenging, as the effect of the variable $a$ can be easily overshadowed by the covariates $\bm{X}$ if the covariate space is high-dimensional \cite{nie_vcnet_2021, wang_generalization_2022, nagalapatti_continuous_2024}. In the time-varying setting we consider in this work, this task is even more difficult, as we have two special variables: $a$ and $t$.

\subsection{What if time measurements are not assigned randomly?}
In case when the treatment is assigned non-randomly, point identification of $\tau_t(a)$ from observational data is possible under the following weaker assumptions \citep{vanderschueren_accounting_2023}:
\begin{assumption}[Sampling at Random (SAR)]
    For all times $t \in \mathcal{T}$ and for all dosage assignments $a \in \mathcal{A}$, the potential outcome $Y_a(t)$ is conditionally independent from the time of measurement, given the covariates $\bm{X}$: $Y_t(a) \perp T \vert \bm{X}$.
\end{assumption}
We further need to make the overlap assumption.
\begin{assumption}[Time Overlap]
    The time of measurement assignment is non-deterministic: there exists a constant $c > 0$ such that $\forall \bm{x} \in \mathcal{X}$ such that $P(\bm{X} = \bm{x}) > 0$, we have $c < P(T = t \vert \bm{X}=\bm{x}) < 1 - c$.
\end{assumption}
In those cases, inverse propensity weighting schemes can be further employed to recover the unbiased estimate of $\tau_t(a)$. See \cite{vanderschueren_accounting_2023, lin_analysis_2004} for an overview.

\section{Additional Experiments}

\subsection{Baseline Trajectory Model Comparison}
\label{appendix: a0_model}
We explore three alternative baseline trajectory models $\hat{\varphi}(\bm{x}, t)$. Firstly, we consider two standard regression models fitted to the concatenated variables $(\bm{x}, t)$ -- a XGBoost model and a neural network model. Although these approaches are straightforward, when $\bm{X}$ is high-dimensional, the time component $t$ may be overshadowed by the multitude of covariates, reducing temporal accuracy. For an alternative strategy, we use the patient covariates $\bm{x}$ to identify the 1-nearest neighbour of patient $i$ in the $\mathcal{X}$ space -- patient $k$. Since patient $k$'s measurement times $\{T_{k, j}\}$ may not align with those of patient $i$, we spline interpolate $\{Y_{k, j}\}$ to estimate $Y_{T_{i, j}}(0)$. This can be suboptimal when the measurement times are heavily misaligned, making spline interpolation unreliable.

\begin{figure}
    \centering
    \includegraphics[width=\linewidth]{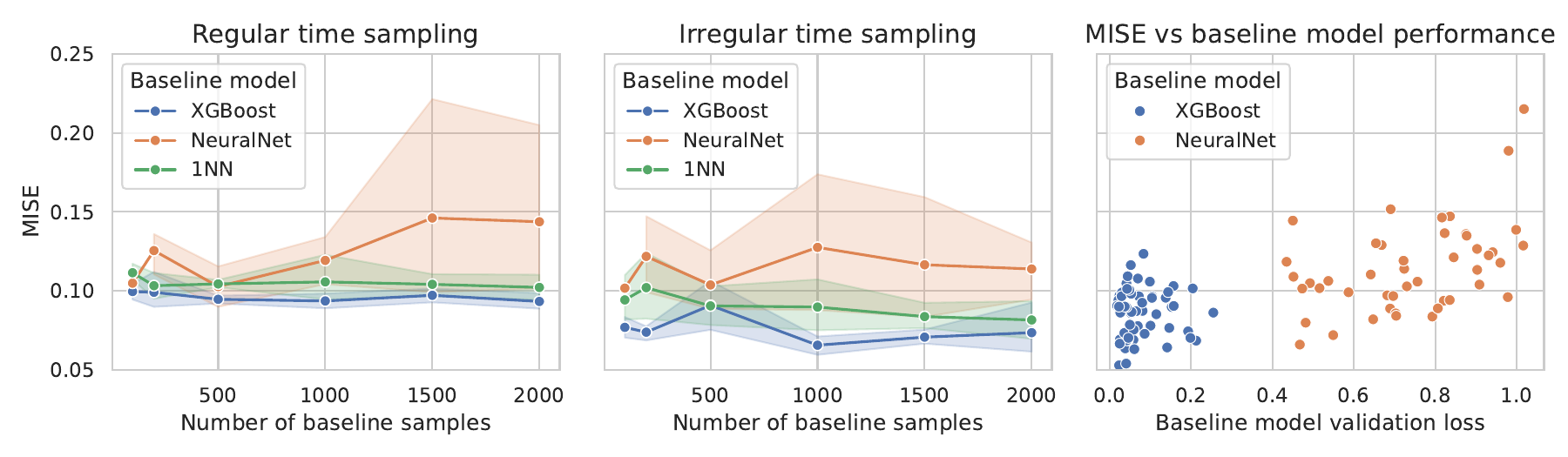}
    \caption{Comparison of the performance of the different baseline trajectory models. The error bars denote the confidence intervals, computed over 10 seeds.}
    \label{fig: baseline_models}
\end{figure}

We compare the performance of SemanticATE, instantiated with different baseline trajectory models, as we vary the number of baseline samples included in the \textit{IHDP-based} dataset, $n_0$. The number of treated patients is kept constant, at $n=200$. The results are visualised in \cref{fig: baseline_models}. 

\paragraph{Results.} The results demonstrate that XGBoost is the most optimal baseline trajectory model in our setting, across all sample sizes. Interestingly, the performance of XGBoost improves as we switch to irregular sampling of the measurement times. As expected, the 1NN method, while competitive in the regular sampling setting, is not capable of as efficiently utilising the information available if the times of measurement are misaligned between the identified nearest neighbours. In the right-hand side plot we showcase a scatterplot visualising the relationship between the validation loss of the baseline trajectory model, calculated after training, and the MISE. The results validate that improving the performance of the baseline trajectory model can significantly improve the $\tau_t(a)$ estimation performance.

\subsection{Sample Efficiency - Treatment Trajectory Model Comparison}
\label{appendix: n_samples}
In here, we further analyse the sample efficiency of SemanticATE, by analysing how the estimation performance depends on the number of \textit{treated} samples (i.e. the number of samples to which the treatment effect trajectory model is fitted). We conduct the experiment on the \textit{IHDP-based} dataset The number of baseline samples is kept constant, at $n_0=1000$. We compare the performance of SemanticODE against the performance of two black-box methods which demonstrated best performance on the IHDP dataset - the XGBoost model and the polynomial regression model.

\begin{figure}
    \centering
    \includegraphics[width=0.4\linewidth]{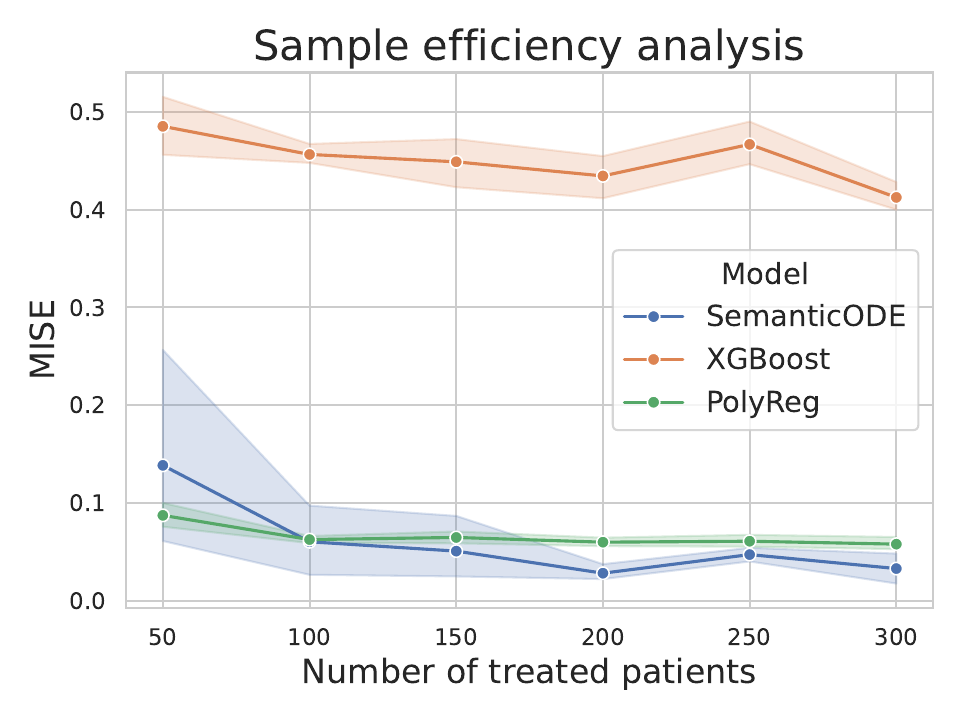}
    \caption{Comparison of the performance of the different trajectory models, as a function of the number of treated patients. The error bars mark the confidence intervals, computed over 5 seeds.}
    \label{fig: n_samples}
\end{figure}

\paragraph{Results.} The results, visualised in \cref{fig: n_samples} demonstrate that SemanticODE, on top of being transparent and easy to verify, is also very sample efficient, with optimal performance, matching the level of the black-box polynomial regression model, reached at the level of as few as 100 samples.

\section{Experimental Design}
\label{appendix: experiments}

The code used to run the experiments was largely based on the \textit{SemanticODE} python library \cite{kacprzyk_no_2024}, available from \url{https://github.com/krzysztof-kacprzyk/SemanticODE}.

\subsection{Details of the Datasets}
\label{appendix: datasets}

\subsubsection{PK Dataset}

\paragraph{Covariates.} In the dataset, the outcome evolution depends on $d=8$ static covariates.

\textit{PK-random.} For this dataset, covariates are sampled uniformly within pre-specified ranges or chosen randomly from discrete (binary) options.

\textit{PK-real.} Synthetic covariates are generated from a multivariate normal distribution, with the mean and covariance matrix estimated from the real data, used in the original study by \cite{woillard_population_2011}. Binary covariates (SEX, CYP, FORM) are thresholded to maintain their discrete nature. This allows us to manipulate the number of samples in the dataset, for a more thorough evaluation of the considered methods.

\textbf{Treatment Assignment.} We firstly assign the baseline dose of $a=0$ to  $n_0$ randomly chosen patients. For the remaining patients, we sample the dose uniformly at random from the range $a_{min}=3$ to $a_{max}=10$.

\textbf{Measurement Time Assignment.} For each patient, unless otherwise stated, we generate $n_t=20$ measurement points, sampled uniformly at random from between 0 and 24. We then order the measurement points in an increasing order.

\textbf{Outcome Trajectories.} The drug concentration over time is modelled using a system of ordinary differential equations (ODEs), identified by \cite{woillard_population_2011}. 

\begin{align*}
\frac{d\,\text{DEPOT}}{dt} &= -KTR \cdot \text{DEPOT} \\
\frac{d\,\text{TRANS1}}{dt} &= KTR \cdot \text{DEPOT} - KTR \cdot \text{TRANS1} \\
\frac{d\,\text{TRANS2}}{dt} &= KTR \cdot \text{TRANS1} - KTR \cdot \text{TRANS2} \\
\frac{d\,\text{TRANS3}}{dt} &= KTR \cdot \text{TRANS2} - KTR \cdot \text{TRANS3} \\
\frac{d\,\text{CENT}}{dt} &= KTR \cdot \text{TRANS3} - \left( \frac{CL + Q}{V1} \right) \cdot \text{CENT} + \frac{Q}{V2} \cdot \text{PERI} \\
\frac{d\,\text{PERI}}{dt} &= \frac{Q}{V1} \cdot \text{CENT} - \frac{Q}{V2} \cdot \text{PERI}
\end{align*}

The values of the parameters used in the model are summarised in the table below.

\begin{table}[]
    \centering
    \begin{tabular}{ll}
    \hline
    Parameter & Value \\
    \hline
    $CL$ & 80.247 \\
    $V1$ & 486 \\
    $Q$ & 79 \\
    $V2$ & 271 \\
    $KTR$ & 3.34 \\
    DEPOT$(t_0)$ & 10 \\
    CENT$(t_0)$ & $y_0 \times (V1 / 1000)$ \\
    PERI$(t_0)$ & $V2 /V1 \times$CENT$(t_0)$ \\
    TRANS1$(t_0)$ & 0.0 \\
    TRANS2$(t_0)$ & 0.0 \\
    TRANS2$(t_0)$ & 0.0 \\
    \hline
    \end{tabular}
    \caption{Parameters of the PK Dataset.}
    \label{tab: pkdataset}
\end{table}

After obtaining the covariates for each patient, we solve the initial value problem to obtain a trajectory of CENT as our outcome. We then scale it back to appropriate units by multiplying by 1000 and dividing by V1. Then, we add Gaussian noise with the standard deviation $\sigma=0.01$ to the scaled outcomes.

\subsubsection{IHDP-based Dataset} 

\paragraph{Covariates.} For this dataset, covariates are generated to resemble the real covariate distributions from the Infant Health and Development Program (IHDP) dataset. Specifically, synthetic covariates are drawn from a multivariate normal distribution, with the mean and covariance matrix estimated from the real IHDP data, obtained from \url{https://github.com/vdorie/npci/raw/master/examples/ihdp_sim/data/ihdp.RData}. Binary covariates are thresholded to maintain their discrete nature. This approach allows for the manipulation of the number of samples in the dataset, facilitating a more thorough evaluation of the considered methods.

\textbf{Treatment Assignment.} We first assign the baseline dose of $a=0$ to $n_0$ randomly chosen patients. For the remaining patients, we sample the dose uniformly at random from the range $a_{min}=2$ to $a_{max}=6$.

\textbf{Measurement Time Assignment.} For each patient, unless otherwise stated, we generate $n_t=20$ measurement time points, sampled uniformly at random from between 0 and 60. These time points are then ordered in increasing sequence.

\textbf{Outcome Trajectories.} The drug concentration over time is modelled using a combination of baseline outcomes and treatment effects, according to the formula:

$$y(t, a, \bm{x}) = \tau_t(a) + h(a, \bm{x}) + y^0_t(\bm{x}) + \epsilon.$$

The  Treatment effects are incorporated through an average treatment effect (ATE) function, which modulates the baseline outcomes according to the administered dose.

\textbf{Baseline Outcome $y^0_t(\bm{x})$:} baseline outcome is generated using B-splines with $n_b = 6$ basis functions, which allow for flexible modelling of time-dependent trajectories based on patient covariates. For each patient, the coefficients of the B-spline are obtained by multiplying the vector of the patient covariates by a set of randomly generated weights.
    
\textbf{Static CATE function $h(a, \bm{x})$:} To introduce treatment effect heterogeneity into the dataset, to each trajectory we add a static treatment effect value. For the definition of $h(a, \bm{x})$, we use the continuous-dose version of IHDP, proposed by \cite{nie_vcnet_2021}. Then, to ensure that $\tau_t(a)$ indeed describes the average treatment effect, at each dose level $a$ we calculate the average treatment effect and subtract it from the value of the HTE.

\textbf{Ground-truth function $\tau_t(a)$.} To define a plausible $\tau_t(a)$ curve, we a trajectory defined using the following equations:
\begin{align}
    t_{max}(a) &= 25/(-2.6) \times (a - 6) + 5 \\
    \tau_{max}(a) &= 10 + 60 \times (1/30 \times t_{max})\times(1-\frac{1}{30}t_{max})^3 \\
    s &= 7 \\
    h &= 10 \\
    \phi(x) &= \exp\left(\frac{-1}{2s^2}(x-t_{max})^2\right)
\end{align}
\begin{equation}
    y_t(a) = 
        \begin{cases}
        \frac{3a}{1 + \exp(- \sqrt{\frac{a}{4}}t - 11.5)} - \frac{a/2}{(1 + e^5)}, & \text{if $t_{max} \geq 30$}\\
        \frac{\phi(x) - \phi(0)}{\phi(t_{max}) - \phi(0)}\tau_{max}\mathds{1}\{t \leq t_{max}\} + \left(h + \phi(t)\frac{\tau_{max} - h}{\phi(t_{max})}\right)\mathds{1}\{t > t_{max}\}, & \text{otherwise}\\
        \end{cases}    
\end{equation}

In particular, the values of the parameters were chosen to allow for smooth transitions between shapes of different trajectories.

We begin by generating baseline outcome trajectory and adding a scalar to it so that $y_{min} = 0$. Then, we scale it to have $y_{max}=1$, add a Gaussian noise with standard deviation $\sigma=0.01$ and scale it back to its original scale. This is to ensure that all patients, both the treated and the non-treated, have noise added at the same level. We then add $c \times (\tau_t(a) + h(a, \bm{x}))$ to the trajectory, where $c$ is chosen in a way to ensure that the range of the baseline outcome trajectory is equal to the range of the combined treatment effect. Finally, we scale the value of the outcome by dividing by $y_max$.

\subsection{Dataset pre-processing}
In case of both datasets, the treatments, $a$ and times, $t$ are normalised to fall between 0 and 1 (so that for time $t$, in-distribution time interval is $t \in [0, 1]$ and then out-of-distribution interval is $t \in [1, 1.25]$). Unless otherwise stated, we use $n_0=1000$ and $n=1200$ samples across all experiments, so that the final composition and property maps are fitted using 200 data points. 

\subsection{Baseline Trajectory Models}

\subsubsection{Regression models}
We fit a model $f: \mathcal{X} \times \mathcal{T} \rightarrow \mathcal{Y}$. Before fitting the models, we divide the entire baseline dataset into 70\% train and 30\% validation sets. All variables are standardised before the model fitting. For each model, we performed hyperparameter tuning with 10 randomly sampled trials (unless otherwise stated).The hyperparameter space is defined in the tables below.

\begin{table}[]
    \centering
    \begin{tabular}{c|c}
    \hline
    Parameter name & Range of values \\
    \hline
       learning rate  & [0.0001, 1.0] \\
        max\_depth & [3, 10] \\
        n\_estimators & \{50, 100, 150, 200\} \\
        subsample & \{0.5, 0.6, 0.7, 0.8, 0.9, 1.0\} \\
        colsample\_bytree & \{0.5, 0.6, 0.7, 0.8, 0.9, 1.0\} \\
    \hline
    \end{tabular}
    \caption{Range of hyperparameters for XGBoost.}
    \label{tab: xgboost}
\end{table}

\begin{table}[]
    \centering
    \begin{tabular}{c|c}
    \hline
    Parameter name & Range of values \\
    \hline
       learning rate  & [0.0001, 1.0] \\
       n\_layers & \{1, 2, 3, 4, 5\} \\
       layer\_size & [8, 128] \\
       dropout\_rate & \{0., 0.1, 0.2, 0.3, 0.4, 0.5\} \\
       weight\_decay & \{0., 0.1, 0.2, 0.3, 0.4, 0.5\} \\
       activation & \{tanh, relu\} \\
    \hline
    \end{tabular}
    \caption{Range of hyperparameters for NeuralNet.}
    \label{tab: neuralnet}
\end{table}

\subsubsection{1 Nearest Neighbour}
We identify the 1 nearest neighbour in the $\mathcal{X}$-space. Then, to interpolate the baseline trajectories in case that the outcome measurement times are misaligned, we use polynomial regression of the baseline trajectory with degree 3.

\subsection{Treatment Effect Trajectory Models}
Before fitting the models, we divide the entire baseline dataset into 70\% train and 30\% validation sets. Unless otherwise stated, we tune each of the benchmark models for 20 trials, and the SemanticODE model for 5 trials for each of the property maps.

\subsubsection{SINDy}
We use SINDy \citep{brunton_discovering_2016} as implemented in PySINDy package \citep{silva_pysindy_2020, kaptanoglu_pysindy_2022}. We pass the variable t as an additional dimension of the trajectory to allow for time-dependent solution (not just autonomous systems). We use the following library of functions:
\begin{equation}
1, \; x,\; t,\; x2,\; xt,\; t2,\; e^x,\; e^t,\; sin(x),\; sin(t),\; cos(x),\; cos(t),\; sin(2x),\; sin(2t),\; cos(2x),\; cos(2t), \; sin(3x), \; cos(3x)
\end{equation}
We use the STLSQ optimiser, and we tune the hyperparameter $\alpha$ with values in range [0.001, 1.0]. We further tune the differentiation method used, by choosing from the following techniques: finite difference, spline, trend filtered, and smoothed finite difference as available in PySINDy. The parameter ranges we consider for each of them are shown in the table below.

\begin{table}[]
    \centering
    \begin{tabular}{c|c}
    \hline
       Method  & Range of hyperparameters \\
       \hline
       finite difference & $k \in \{1, \dots, 5\}$ \\
       spline & $s \in (0.001, 1)$ \\
       trend filtered & order $\in \{0, 1, 2\}$, $\alpha \in (0.0001, 1)$ \\
       smoothed finite difference & window\_length $\in \{1, \dots, 5\}$ \\
       \hline
    \end{tabular}
    \caption{Range of the hyperparameters for each of the differentiation methods for SINDy.}
    \label{tab: sindy}
\end{table}

\subsubsection{WSINDy}
We use WSINDy \citep{reinbold_using_2020, messenger_weak_2021} as implemented in PySINDy package \citep{silva_pysindy_2020, kaptanoglu_pysindy_2022}. We pass the variable t as an additional dimension of the trajectory to allow for time-dependent solution (not just autonomous systems). We use the following library of functions:
\begin{equation}
1, \; x,\; t,\; x2,\; xt,\; t2,\; e^x,\; e^t,\; sin(x),\; sin(t),\; cos(x),\; cos(t),\; sin(2x),\; sin(2t),\; cos(2x),\; cos(2t), \; sin(3x), \; cos(3x)
\end{equation}
We use the SR3 optimiser, and we tune the optimizer threshold with values in range [0.001, 1.0].

\subsection{NeuralODE}
We implement the NeuralODE \cite{chen_neural_2018} method in Pytorch. We tune the same hyperparameters and hyperparameter ranges as in the case of baseline trajectory model (see \cref{tab: neuralnet}).

\subsection{XGBoost}
We use the XGBRegressor, as implemented by \cite{chen_xgboost_2016}. We tune the same hyperparameters and hyperparameter ranges as in the case of baseline trajectory model (see \cref{tab: xgboost}).

\subsection{Polynomial Regression}
We use the ridge regression as implemented in scikit-learn. We tune the hyperparameter alpha of the ridge regression, with values between (0.001, 0.0), and the degree of the features for the polynomial expansion, with values between 1 and 5.

\subsection{SemanticODE}
We use the SemanticODE method \cite{kacprzyk_no_2024}, as implemented by \cite{kacprzyk_no_2024}. In all experiments, unless otherwise stated, we use a composition library containing all compositions up to 4 motifs, constrained to end in a horizontal asymptote. We choose the maximum number of branches for the composition map to be $I=3$. Each univariate function in the property maps is described as a linear combination of 6 basis functions: constant, linear, and four B-Spline basis functions of degree 3. The property maps are trained using L-BFGS as implemented in PyTorch. We fix the penalty term for the difference between derivatives to be 0.05 and we perform hyperparameter tuning of each property sub-map to find the optimal learning rate (between $10^{-4}$ and 1.0) and the penalty term for the first derivative at the last transition point (between $10^{-9}$ and $0.1$).


\end{document}